# Risk factor identification and classification of malnutrition among under-five children in Bangladesh: Machine learning and statistical approach

Tasfin Mahmud*, Tayab Uddin Wara, Chironjeet Das Joy

BRAC University, Merul Badda, Dhaka 1212, Bangladesh

**Abstract**

This study aims to understand the factors that resulted in under-five children's malnutrition from the Multiple Indicator Cluster (MICS-2019) nationwide surveys and classify different malnutrition stages based on the four well-established machine learning algorithms, namely- Decision Tree (DT), Random Forest (RF), Support Vector Machine (SVM), and Multi-layer Perceptron (MLP) neural network. Accuracy, precision, recall, and F1 scores are obtained to evaluate the performance of each model. The statistical Pearson correlation coefficient analysis is also done to understand the significant factors related to a child's malnutrition. The eligible data sample for analysis was 21,858 among 24,686 samples from the dataset. Satisfactory and insightful results were obtained in each case and, the RF and MLP performed extraordinarily well. For RF, the accuracy was 98.55%, average precision 98.3%, recall value 95.68%, and F1 score 97.13%. For MLP, the accuracy was 98.69%, average precision 97.62%, recall 90.96%, and F1 score of 97.39%. From the Pearson co-efficient, all negative correlation results are enlisted, and the most significant impacts are found for the WAZ2 (Weight for age Z score WHO) (-0.828''), WHZ2 (Weight for height Z score WHO) (-0.706''), ZBMI (BMI Z score WHO) (-0.656''), BD3 (whether child is still being breastfed) (-0.59''), HAZ2 (Height for age Z score WHO) (-0.452''), CA1 (whether child had diarrhea in last 2 weeks) (-0.34''), Windex5 (Wealth index quantile) (-0.161''), melevel (Mother's education) (-0.132''), and CA14/CA16/CA17 (whether child had illness with fever, cough, and breathing) (-0.04) in successive order.

## 1. Introduction

The prevalence of malnutrition among the Bangladeshi lagged-behind population is vivid, and among the children, this problem is acute. Although the child mortality rate has been declining over the past few decades, it still holds the under-5 death toll of a whopping 27.3 per 1000 births in Bangladesh, according to the UNICEF dataset 2023. Any discrepancies in consuming the micro- and macro-nutrient elements in the daily food chain are referred to as malnutrition [1]. Malnutrition prevalence in the body of children has both short- and long-term consequences, including mortality, disability, morbidity, infection with various diseases, intellectual handicaps, and economic failure [2]. Even severely malnourished children have been experiencing a greater risk of mortality (generally 2 to 4 folds greater than well- and moderately nourished and mildly malnourished children) [3]. A nutrition deficiency in the human body tends to make it vulnerable to infectious diseases, and it debilitates the body to a certain stage that disables the resilience capability to be infected by diseases. Not only the life expectancy ragged to an alarming rate due to the deficiency of nutrition, but the productivity of an entire working-class population decreases in the long run. Therefore, scrutinizing and studying the prevalence of malnutrition in the current scenario is crucial to fostering national growth, as addressing the issue widens the path to resolution.

There are numerous ways to determine a child's nutrition status, among which the traditional method of assessing malnutrition using the Z score has been proven strong and conventional [4]. But it doesn't allow us to delve into the relationship between the other related factors with malnutrition. As a result, surveyors and nutritionists have traditionally followed the statistical approach. Only a handful of studies [5, 6, 7, 8, 9, 10] have been done on analyzing bulk data on malnutrition specifically in Bangladesh using the machine learning algorithm. Almost all of the conducted research is based on the past MICS and BDHS (2014) datasets, but in this research, the latest MICS (2019) dataset was used, which was led by the BBS (Bangladesh Bureau of Statistics) with the help of Unicef. Also, some insightful findings were carried out to classify the different stages of malnutrition in Ethiopia, Zambia, Afghanistan, Yemen, Philippines, India, Papua New Guinea, and sub-Saharan African countries [11, 12, 13, 14, 15, 16, 17, 18]. No work has been done with a high degree of accuracy, utilizing and preparing the datasets in the most sophisticated way to feed the machine learning models along with the statistical analysis. To mitigate these research gaps, four machine learning algorithms—i) Decision Tree (DT), ii) Random Forest (RF), iii) Support Vector Machine (SVM), and iv) Multi-layer Perceptron (MLP)—were used. Also, a statistical approach using a negative Pearson correlation coefficient was obtained to solidify the research findings and understand the risk factors that are associated with child malnutrition. A systematic assessment is also done of accuracy, precision, recall, and the F1 score associated with each of the machine learning approaches as performance metrics.

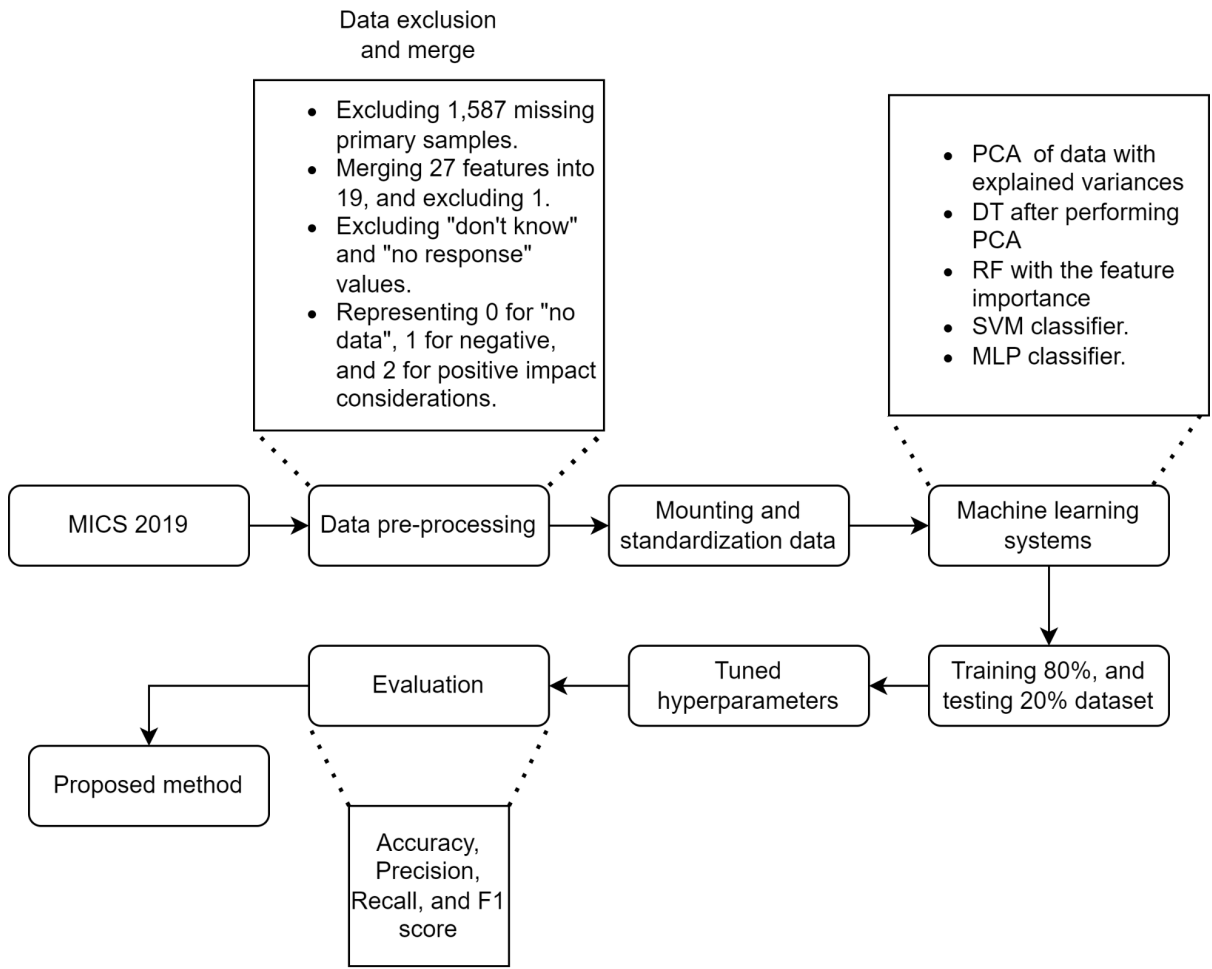

Fig. 1. Complete workflow diagram for the machine learning algorithm-based approach

## 2. Materials and methods

Fig. 1 depicts the overall workflow of the machine-learning approaches. In this work, the secondary collected data MICS (2019) survey was used as the input dataset. After processing the dataset, it was ready to be fed to the ML algorithms. The principal components of the dataset were stratified by performing PCA, which the algorithm itself processes to form interrelated features. The impactful risk factors were obtained from the DT and RF algorithms, as well as the SVM and MLP for predicting malnutrition. On the other hand, through the statistical Pearson correlation, an insightful relationship between malnutrition and its impacted features was observed.

*2.1 Data and variables*

The intensive survey conducted across Bangladesh from January 19, 2019, to June 1, 2019, the Multiple Indicator Cluster Survey (MICS), which was conducted by Unicef, BBS, and partially funded by the UNFPA, was applied in this analysis. In total 61,242 households were interviewed during the survey with a total of 144 significant indicators related to children and women acquired, and among those, almost 29 indicators were directly connected to the Sustainable Development Goals (SDGs). It represents the national averages on those significant indicators and a benchmark for the different socio-economic axes based on reliable data from seven divisions' 64 districts [19]. From the Bangladesh MICS SPSS datasets, the "ch" unit, which was interviewed from the perspective of "Mothers or primary caretakers of children under the age of five," was included in this analysis. The total collected samples were counted at 24,686 in this unit, but within those 21,858 samples were finally analyzed. Initially, due to the missing 1,587 primary components of the datasets, like the BMI Z score WHO (ZBMI), weight for age Z score WHO (WAZ2), weight for height Z score WHO (WHZ2), and height for age Z score WHO (HAZ2), the entire sample got redundant.

From the surveyed information, diarrhea and pneumonia-related features were categorized and merged at first. Then, excluding the district indicators due to the large variances, the primary component features of the Z scores were classified as the malnutrition status (Mstatus) into four segments, namely: well nourished (-1< Z score < 0, and up), mildly malnourished (-2< Z score<-1), moderately malnourished (-3< Z score<-2), and severely malnourished (Z score<-3) based on the WHO standard and represented as 1, 2, 3, and 4, respectively. Additionally, the other features were extracted and stated mostly as 0, 1, and 2 so that the computation of the machine learning algorithms would be the most efficient. Among these, 0 was for no information, 1 for all the negatively correlated factors, and 2 for all the positively correlated terms. But the age of the child (UB2) has continuous values, and the wealth index quintile (Windex5) and the mother's education level (melevel) have different standards ranging from low (1, and 0) to high (5, and 3), respectively. These stratifications of the dataset had a crucial role to play in each approach.

**Table 1**
Statistical information on some of the significant features of the dataset

| | Feature names | | | | | | | |
|---|---|---|---|---|---|---|---|---|
| | UB2 | BD3 | Windex5 | melevel | ZBMI | WHZ2 | WAZ2 | HAZ2 |
| Data type | contin. | categor. | categor. | categor. | contin. | contin. | contin. | contin. |
| Statistical quantity | | | | | | | | |
| Minimum | 0.00 | 0.00 | 1.00 | 0.00 | -4.99 | -4.99 | -5.66 | -6.00 |
| Maximum | 4.00 | 2.00 | 5.00 | 3.00 | 4.98 | 4.86 | 4.68 | 5.86 |
| Mean | 2.02 | 1.03 | 2.81 | 1.69 | -0.57 | -0.66 | -1.21 | -1.29 |
| Standard Deviation | 1.41 | 0.95 | 1.42 | 0.86 | 1.17 | 1.16 | 1.10 | 1.32 |

Table 1 illustrates the different statistical parameters, such as the minimum, maximum, mean, and standard deviation values of some of the important variables.

Apart from these significant features, the other relevant features (all categorical) that have been considered in this work are depicted in Table 2.

**Table 2**
Description of the rest of the features of the dataset

| Feature name | Description |
| --- | --- |
| CA1 | Whether child had diarrhea in last 2 weeks |
| CA3 | Whether child drank less or more during diarrhea |
| CA4 | Whether child ate less or more during diarrhea |
| CA5 | Whether sought advice or treatment for the diarrhea from any source |
| CA7A/CA7B | Whether child had saline during diarrhea |
| CA7C/CA13B/CA13G | Whether child had medications during diarrhea |
| CA14/CA16/CA17 | Whether child had illness with fever, cough, and breathing |
| CA22/CA23L/CA23M/CA23N | Whether child had medications during cough |
| HH6 | Area of living |
| HH7 | Division of the respondent |
| HL4 | Gender of the child |

## 2.2 Machine learning models

### Principal Component Analysis (PCA)

Principal Component Analysis (PCA) is a technique for dimensionality reduction of a dataset which is widely used in machine learning and statistics. While preserving as much as the original value, PCA's main goal is to reduce the high dimensionality of a dataset into a lower-dimensional representation so that the computation cost can be lower and the data analysis and the modeling processes can be simplified. However, since PCA works on assuming that there is a linear relationship between variables while using non-linear dataset, the effectiveness of PCA may be limited. Also if dataset consists of numerous variables, PCA may not give a clear or intuitive interpretation.

For $N$-dimensional input data, it is projected on a $k$-dimensional linear subspace ensuring the minimization of the reconstruction error. Sum of the squared of the $L_2$-distances between original and projected data is considered as the error for this minimization process. If $X$ is the input data matrix that satisfies the condition to be mean-centered (i.e. $\sum_{i=1}^{i=m} x_i = 0$), $P_k$ is defined as the set of rank-$k$ orthogonal projection matrices having $N$-dimension. PCA is then defined as the following minimization problem by the orthogonal projection matrix solution $P^*$:

$$\min_{P \in P_k} \| PX - X \|_F^2$$

The PCA solution, $P^* \in P_k$ can be determined using the following equation:

$$P^* = U_k U_k^T$$

where $U_k \in \mathbb{R}^{N \times k}$ is the matrix formed by the top $k$ singular vectors of $C = \frac{1}{m} XX^T$ and $C$ is the sample covariance matrix corresponding to $X$.

Total 20 features were extracted from the MICS – 2019 raw dataset in this work. To reduce the number of features, PCA was done before feeding the dataset in Decision Tree algorithm to ensure better accuracy. At the time of performing PCA 95% variance was kept, so that the dataset does not lose any significant information. After doing so, 11 principle components were obtained which have been portrayed further in the results section.

*Decision Tree (DT)*

Among all the supervised learning methods Decision Tree (DT) is used for both classification and regression tasks, it is exactly what its name suggest, a tree-like model where an internal mode represents a feature or variable, the branch represents a decision rule and the label or outcomes is represented by each leaf node. Due to its simplicity interpretability and versatility, DT is widely used. However, it can produce overfitted results if not the pruning done carefully. Also it may show instability as small change in the data might produce different tree structure. Complex linkages in the data may not be as well captured by a single decision tree as they are by more sophisticated models.

For effectively classifying different malnutrition status using the Decision Tree, entropy was selected in this work for measuring the information content:

$$\text{Info}(X) = -\sum_{i=1}^{K} p_i \log_2 p_i$$

where, $\text{Info}(X)$ corresponds to the average information for identifying the label of a class of a tuple $X$, with the probability $p_i$ of individual classes. For this work, the total number of classes, $K = 4$.

The expected information has been calculated from the tuple $X$ based on the partitioning by A:

$$\text{Info}_A(X) = \sum_{j=1}^{V} \frac{|X_j|}{|X|} \times \text{Info}(X_j)$$

Here, $\frac{|X_j|}{|X|}$ represents the weight of the $j^{th}$ partition. Finally, the highest information gain was determined for separating characteristics with attribute $A$, at any node:

$$\text{Gain}(A) = \text{Info}(X) - \text{Info}_A(X)$$

Among the 11 principle components fed to the Decision Tree algorithm, 8 had significant weight to classify the malnutrition status. The resultant decision tree had the maximum depth of 17, with a total number of 1479 nodes.

*Random Forest (RF)*

Random forest is one of the ensemble learning methods that is used in both regression and classification processes. By using random features and bootstrapping sampling, it creates several decision trees to observe the variation among them. In the process, every tree produces individual outcomes and by using average or majority voting, the decision is taken. As the RF is an ensemble learning method, it has excellent accuracy and is not prone to overfitting. Also, while producing the outcomes, it calculates the significance of features, which is very helpful for variable selection.

In a random forest, multiple trees are grown independently and their results are added by a voting mechanism for regression and classification. The ensemble indication F(x) is determined by aggregating each prediction $\hat{f}^{*b}(x)$. The Bagging estimate is defined by [20]:

$$\hat{f}_{bag}(x) = \frac{1}{B}\sum_{b=1}^{B}\hat{f}^{*b}(x)$$

where, $\hat{f}_{bag}(x) \rightarrow \hat{f}(x)$ as $B \rightarrow \infty$.

To determine the maximum number of estimators for Random Forest classifier, out-of-bag (OOB) error was calculated to observe whether it becomes steady. OOB error can be defined as:

$$\text{OOB Error} = \frac{1}{N}\sum_{i=1}^{N}l(y_i \neq \hat{y}_{OOB,i})$$

where, $N$ is the number of training data points, $y_i$ is the actual label, and $\hat{y}_{OOB,i}$ is the aggregated OOB prediction of the $i^{th}$ data point, respectively. $l(.)$ represents the indicator function, which is 1 if the argument is true and 0 otherwise.

In Fig. 2, the OOB error rate seems to be stable for almost 80 estimators. For this reason, the maximum number of estimators was explicitly limited to 100 in this work, to reduce computational cost, yet to attain better accuracy.

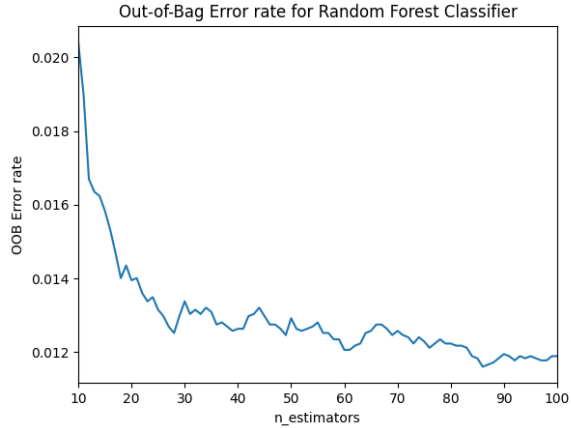

Fig. 2. OOB Error rate for Random Forest classifier to determine number of estimators

*Support Vector Machine (SVM)*

To solve regression and classification problems, a support vector machine is used to find the best hyperplane in a high dimensional space to optimize the edges between classes. It can handle non-linearly separable data through the kernel approach. Even though SVM is great at text and image classification, it is more sensitive to noise, which may affect its accuracy and it requires more computation power.

$N$ pairs of training data consists in the training set $(x_1, y_1), (x_2, y_2)...., (x_N, y_N)$ with $x_i \in R^P$ and $y_i \in \{1,2,3,4\}$. A hyperplane is defined by-

$$\{x : f(x) = x^T\beta + \beta_0 = 0\}$$

Where $\beta$ is defined by a unit vector: $\|\beta\| = 1$. And, $f(x)$ is included by a rule of classification:

$$G(x) = sign[x^T \beta + \beta_0]$$

In this work, the Radial Basis Function (RBF) kernel was used to compute the similarity of two points. For two points, $P_1$ and $P_2$, the RBF kernel can be represented as follows:

$$K(P_1, P_2) = \exp(-\frac{\|P_1 - P_2\|^2}{2\sigma^2})$$

where, $\sigma$ is the variance, and has been considered as one of the hyperparameters for training the model, and $\|P_1 - P_2\|$ is the $L_2$ norm of the two points, $P_1$ and $P_2$.

*Multi-layer Perceptron (MLP)*

Multi-layer perceptron is a Neural Network architecture that consists of input layers, hidden layers, and output layers. It is a feedforward mechanism that processes any data by feeding it into perceptrons, which assign different weights to each variable. To utilize complex patterns in the data, different activation functions, such as $ReLU$ and the sigmoid function, are used. MLP uses iterative processes to adjust the weights and bias, which is called backpropagation so that it can minimize the difference between predicted output and actual output. Generally, activation functions such as $softmax$ and $ReLU$ are used. The $softmax$ is represented as such:

$$softmax(x_i) = \frac{e^{x_i}}{\sum_{i=1}^{K} e^{x_i}}$$

where, $K$ is the number of classes, and

$$ReLU(x) = \max(0, x)$$

With the above activation functions, MLP can make predictions using the equation below [21]:

$$\hat{y} = activation(\sum w_i x_i + b)$$

where, $\hat{y}$ is predicted output, $b$ is the bias and $w_i$ is the weight of the corresponding input vectors, $x_i$.

Three activation layers of $ReLU$ were added to this neural network model to initiate non-linearity and make it better at predicting or categorizing practical situations. Additionally, a $softmax$ layer has the ability to interpret sophisticated probability distributions in classification problems. Merging two of these activation functions resulted in a better classification approach for the four malnutrition classes. The detailed architecture of the MLP network has been provided in the results section.

*2.3 Statistical approach*

As the main objective of this work was to identify the probable risk factors of a malnourished child, a Pearson correlation coefficient was performed to understand the impacted features. A Pearson correlation coefficient is a statistical approach quantifying the linear relationship between two opted variables. A range of values between -1 to

1, commencing a level of the most negatively to positively significant correlation within the two variables. A weak, moderate, and strong correlation variation exists for the modulus scores 0.1 < 0.3, 0.3 < 0.5, and > 0.5, respectively. The closer the values are to -1, and 1, the stronger the predictor variable's influence on the dependent predicted variable.

All the pre-processed samples from Excel were fed to the IBM SPSS version 22.0 software for further analysis and processing. Here, the targeted object was the malnutrition status (Mstatus), which was obtained and classified accordingly. And all the general featured coefficients were compiled during this bivariate, 2-tailed statistically significant test. However, only the negatively significant correlations prevailed in Table 3, as the more negative correlations there, the more malnourished a child would be according to the increased Mstatus order (from 1 to 4) of malnutrition.

## 3. Result

*3.1 Decision Tree with PCA*

PCA was performed as the number of features was large considering our dataset. To ensure 95% variance for the principal components, we needed 11 principal components. In Fig. 3, the contribution of each of the principal components to the cumulative variance is shown.

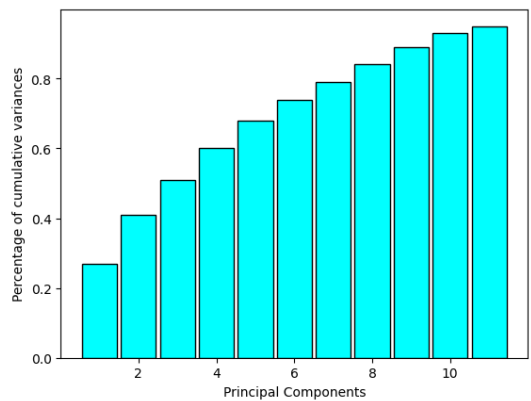

Fig. 3. Cumulative Variance for selected 11 principal components

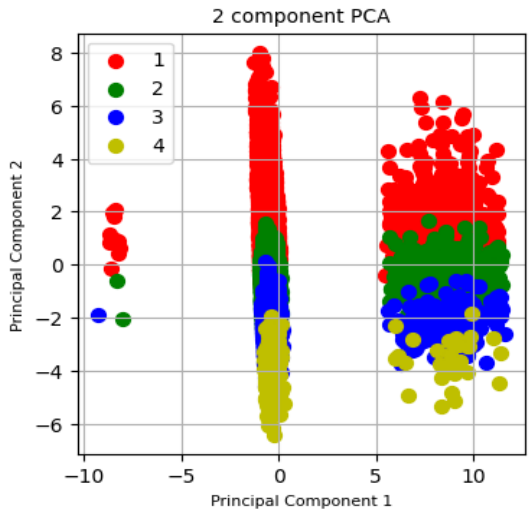
(A) Principal Comp. 1 vs. Principal Comp. 2

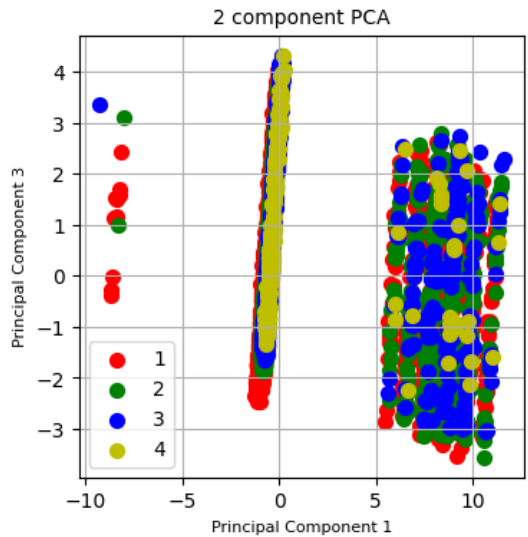
(B) Principal Comp. 1 vs. Principal Comp. 3

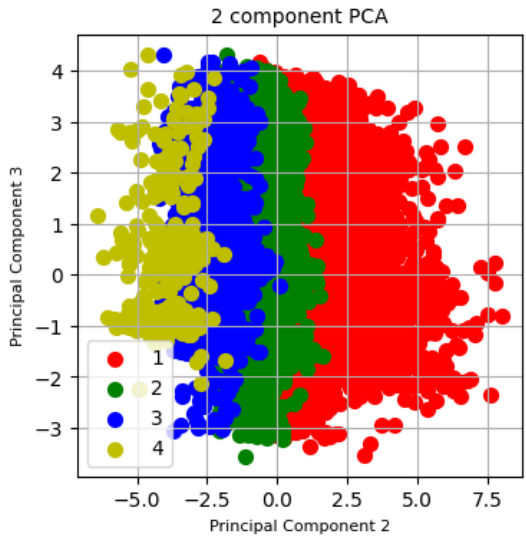
(C) Principal Comp. 2 vs. Principal Comp. 3

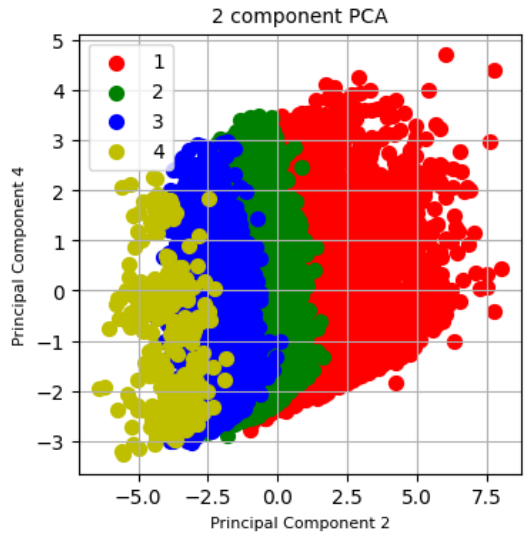
(D) Principal Comp. 2 vs. Principal Comp. 4

Fig. 4. Volcano plots for different Principal components. (A) Principal Comp. 1 vs. Principal Comp. 2; (B) Principal Comp. 1 vs. Principal Comp. 3; (C) Principal Comp. 2 vs. Principal Comp. 3; (D) Principal Comp. 2 vs. Principal Comp. 4

Fig. 4 depicts the principal components among the randomly selected components (between 1 & 2, 1 & 3, 2 & 3, and 2 & 4) and the scattering of the analyzed components was tried to be visualized. It shows how the considered components are independent from each other. Additionally, Fig. 5 shows the confusion matrix for the Decision Tree algorithm.

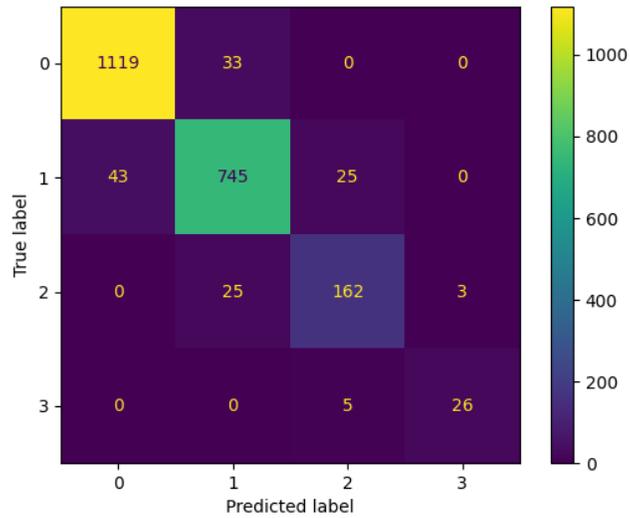

Fig. 5. Confusion matrix for Decision Tree after performing PCA

*3.2 Random Forest (RF)*

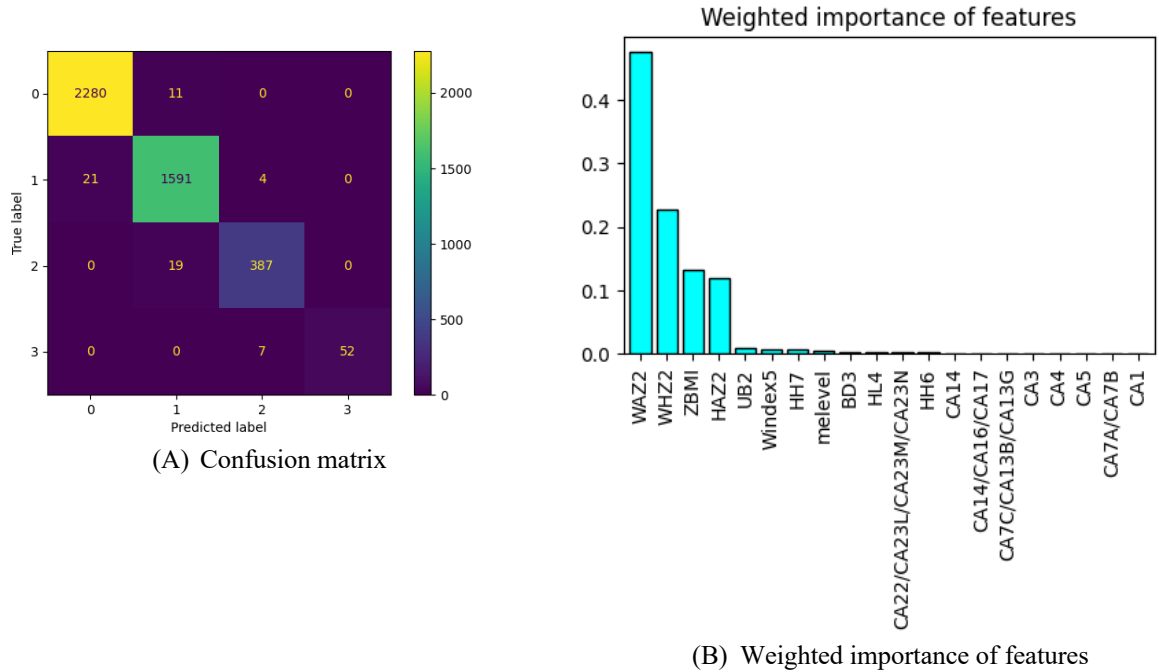

(A) Confusion matrix

(B) Weighted importance of features

Fig. 6. Results from Random Forest. (A) Confusion matrix; (B) Weighted importance of features

Fig. 6 is composed of two segments, one is the confusion matrix obtained from the RF, and the other is the important features that are most influential on the occurrence of child malnutrition, displayed in a sequential and dominant order.

*3.3 Support Vector Machine (SVM)*

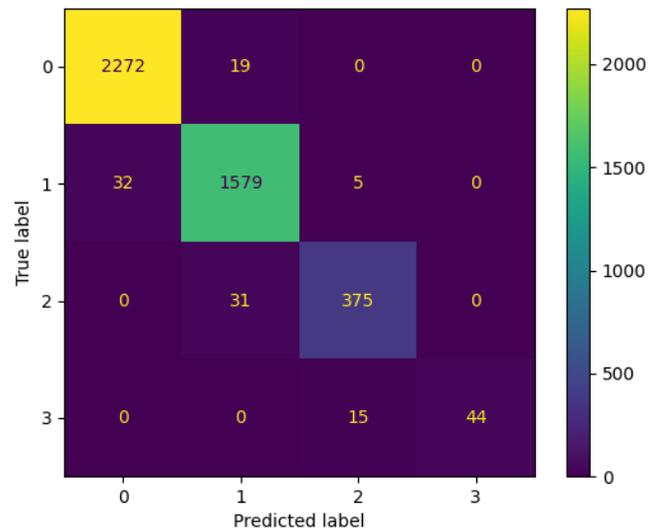

Fig. 7. Confusion matrix for Support Vector Machine

The instances proportion in Fig. 7 shows a slight deviated accuracy of predicting each class in SVM than in Fig. 6 which was for RF.

*3.4 Multi-layer Perceptron (MLP)*

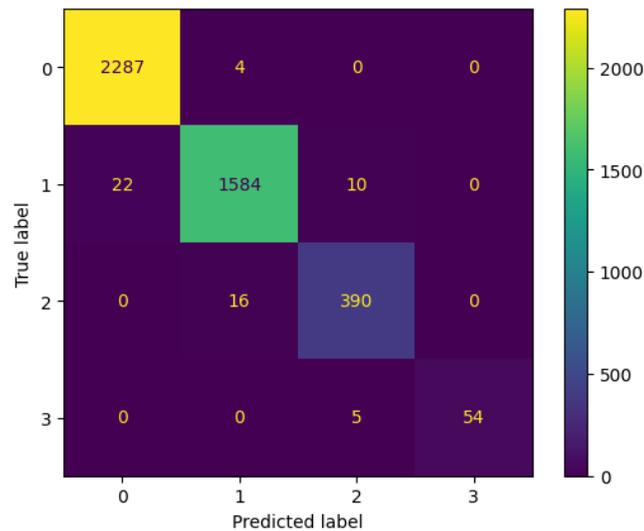

Fig. 8. Confusion matrix for Multi-layer Perceptron

MLP performed well in each classification level, which is portrayed in Fig. 8. As this model could be highly optimized, its accuracy and performance were higher.

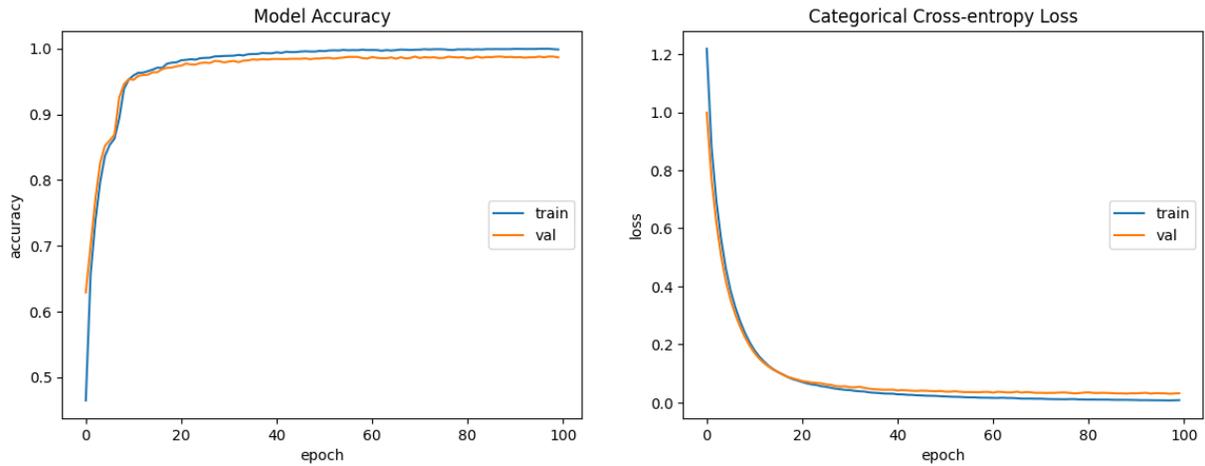

(A) Accuracy vs. epoch  (B) Loss vs. epoch

Fig. 9. (A) Accuracy curve for training and validation split. (B) Categorical loss curve for training and validation splits.

In a window of 100 epochs, the accuracy and categorical cross-entropy losses are plotted for both the training and validation sets in Fig 9. The level of saturation has been obtained in each case, and for accuracy, the training set has a higher degree. On the other hand, the losses are greater than in the validation samples.

**Table 3**

| Type | Output Shape | Parameter |
| --- | --- | --- |
| Dense | (100, ) | 2100 |
| Dense | (40, ) | 4040 |
| Dense | (20, ) | 820 |
| Dense | (4, ) | 84 |

Total params: 7044 (27.52 KB)
Trainable params: 7044 (27.52 KB)
Non-trainable params: 0 (0.00 Byte)

The sequential model architecture of the MLP model, with a total of 7044 parameters, was made up of four hidden layers. Table 3 shows the segmentation of each layer's parameters.

**Table 4**
Performance of the considered ML models

|  | Algorithms | | | |
|---|---|---|---|---|
|  | DT with PCA | RF | SVM | MLP |
| Accuracy (%) (95% CI) | 93.87% | 98.58% | 97.57% | 98.88% |
| Precision (%) (Avg) | 90.25% | 98.50% | 97.75% | 97.75% |
| Recall (%) (Avg) | 89.50% | 95.00% | 91.00% | 96.75% |
| F1 score (%) (Avg) | 90.25% | 96.75% | 93.75% | 97.50% |
| MCC (%) (Avg) | 86.72% | 96.39% | 93.15% | 96.71% |
| CSI rate (%) (Avg) | 78.63% | 93.98% | 88.58% | 95.58% |
| Error rate (%) (Avg) | 6.13% | 1.42% | 2.43% | 1.12% |

The overall accuracy of each ML model with the other average performance metrics - precision, recall, and f1 score, Matthews correlation coefficient (MCC), Classifier success index (CSI) rate, Error rate is shown in Table 4. The accuracy was calculated using 95% confidence intervals (CI), and for the other matrices, class-wise values were averaged.

$$Accuracy = \frac{T_P + T_N}{T_P + T_N + F_P + F_N}$$

$$Precision = \frac{T_P}{T_P + F_P}$$

$$Recall = \frac{T_P}{T_P + F_N}$$

$$F1 - score = 2 \times \frac{Precision \times Recall}{Precision + Recall}$$

$$MCC = \frac{(T_P \times T_N) - (F_P \times F_N)}{\sqrt{(T_P + F_P) \times (T_P + F_N) \times (T_N + F_P) \times (T_N + F_N)}}$$

$$CSI\ rate = Precsion + Recall - 1$$

$$Error\ rate = \frac{F_P + F_N}{T_P + T_N + F_P + F_N}$$

$T_P$, $T_N$, $F_P$, $F_N$ representing "True Positive", "True Negative", "False Positive", and "False Negative" respectively in the squared confusion matrices labeled between the predicted and true classes.

**Table 5**
Bivariate Pearson correlation coefficients with 2-tailed significance

|  |  | BD3 | CA1 | CA14 CA16 CA17 | ZBMI | HAZ2 | WAZ2 | WHZ2 | Windex 5 | melevel | Mstaus |
|---|---|---|---|---|---|---|---|---|---|---|---|
| Mstatus | Pearson correlation | -0.59'' | -0.34'' | -0.04 | -.656'' | -.452'' | -.828'' | -.706'' | -.161'' | -.132'' | 1 |
|  | Sig. (2-tailed) | .000 | .000 | .558 | .000 | ..000 | .000 | .000 | .000 | .000 |  |
|  | N | 21858 | 21858 | 21858 | 21858 | 21858 | 21858 | 21858 | 21858 | 21858 | 21858 |

**.Correlation is significant at the 0.01 level (2-tailed).

*.Correlation is significant at the 0.05 level (2-tailed).

The influential risk factors of a child malnutrition are analyzed through the Pearson correlation co-efficient considering the target variable Mstatus (malnutrition status representing the four classes of malnutrition severity). Only the negatively significant features were enlisted, in Table 5 as they had the most contribution on rising the level of malnutrition. Here, among the nine negatively significant parameters, BD3, ZBMI, WAZ2, and WHZ2 retained the most strongly, CA1, and HAZ2 moderately, as well as Windex5, and melevel, a weakly influential role in accelerating the risk of malnutrition.

**Discussion**

In this study, the probable risk factors and four classes of malnutrition stages (1. well-nourished, 2. mildly nourished, 3. moderately malnourished, and 4. severely malnourished) among under-five children were classified using five ML-based algorithms, namely DT with PCA, RF, SVM, and MLP-based on the MICS 2019 database. To add to that, a bivariate, 2-tailed Pearson correlation coefficient was performed targeting the feature of malnutrition status. In total, 21,858 samples were fed into the ML algorithms as well as into the statistical model. From the extracted 18 features, only 11 components were analyzed in PCA. In Fig. 3, cumulative PCA components and their contribution were depicted, which tends to be gradually ascending. In Fig. 4, the volcano maps of four randomly selected attributes were plotted to visualize the scatteredness of the intended components within those. After performing the PCA analysis, the DT was executed with a classifying accuracy of 93.87%, precision of the four classes 96.29%, 92.78%, 84.38%, and 89.66%, respectively, recall value of the four stages 97.14%, 91.64%, 85.26%, and 83.87%, respectively, and f1 score of each class 96.72%, 92.20%, 84.82%, and 86.67%, respectively, and MCC of each class 92.48%, 87.22%, 83.53%, 83.64%, respectively, and CSI rate of each class 92.93%, 83.84%, 69.99%, 67.74%, respectively, depicted in Fig. 10. For the RF execution, accuracy was 98.58%, precision for the four classes was 99.09%, 98.15%, 97.24%, and 100%, respectively, and recall value for each class was 99.52%, 98.45%, 95.32%, and 88.14%, respectively, and the f1 score of all stages was 99.30%, 98.30%, 96.27%, and 93.69%, respectively, and MCC of each class 98.53%, 97.3%, 95.9%, 93.8%, respectively, and CSI rate of each class 98.6%, 96.6%, 92.57%, 88.14%, respectively. From Fig. 6(B), the weighted importance of features can be shown where WAZ2, WHZ2, ZBMI, and HAZ2 hold the most significance as the key risk factors behind malnutrition. The other eight less significant features were UB2, Windex5, HH7, melevel, BD3, HL4, CA22/CA23L/CA23M/CA23N, and HH6, and the rest of the eight features do not influence malnutrition. Then SVM was performed with an accuracy of 97.67%, precision for four classes was 98.61%, 96.93%, 94.94%, and 100%, respectively, recall values for each stage were 99.17%, 97.71%, 92.36%, and 74.58%, respectively, and the f1 score of each class was 98.89%, 97.32%, 93.63%, and 85.44%, respectively, and MCC of each class 97.66%, 95.74%, 93.0%, 86.2%, respectively, and CSI rate of each class 97.78%, 94.64%, 87.3%, 74.58%, respectively. Lastly, the MLP consisted of four hidden layers (100, 40, 20, and 4) with the relu and softmax activation, and adam solver performed the highest degree of accuracy of 98.69%, precision for the four classes was 99.05%, 98.75%, 96.29%, and 100%, respectively; recall values of each stage were 99.83%, 98.02%, 96.06%, and 91.53%, respectively; and the f1 scores were 99.43%, 98.39%, 96.18%, and 95.58%, respectively, and MCC of each class 98.58%, 97.1%, 95.53%, 95.63%, respectively, and CSI rate of each class 98.65%, 96.34%, 91.9%, 91.43%, respectively on the Keras platform.

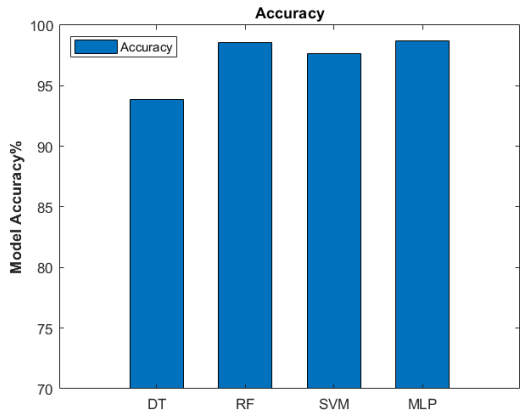

(A) Accuracy for different algorithms

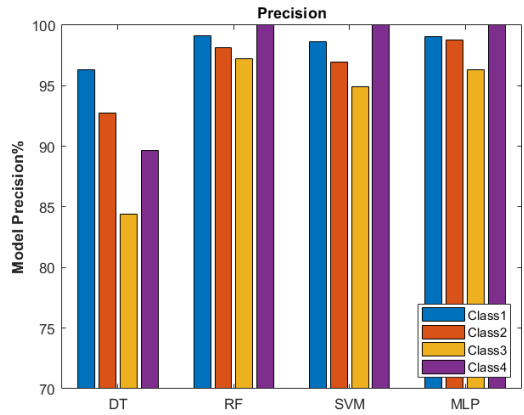

(B) Precision of 4 classes for different algorithms

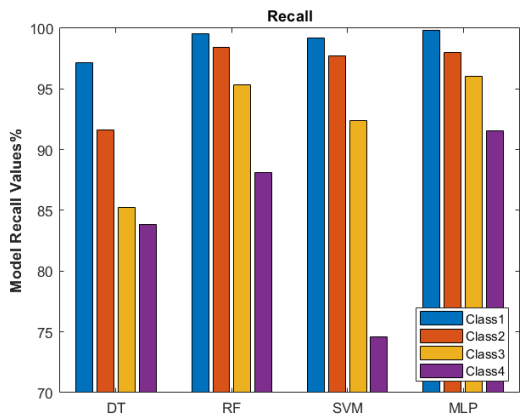

(C) Recall of 4 classes for different algorithms

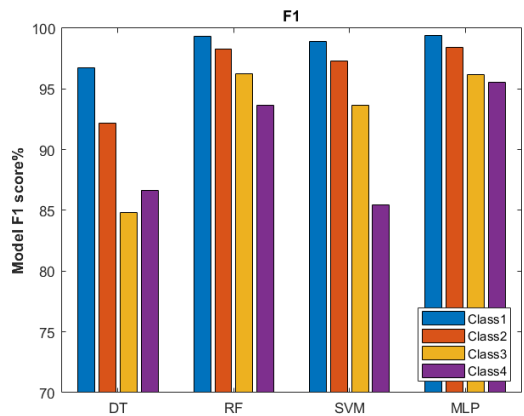

(D) F1-score of 4 classes for different algorithms

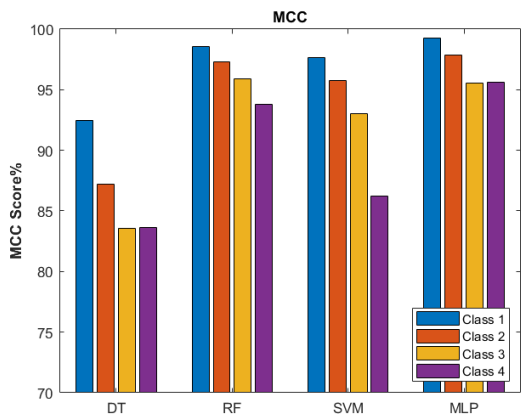

(E) MCC of 4 classes for different algorithms

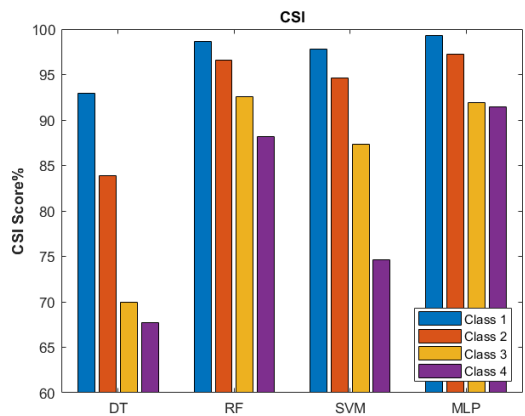

(F) CSI rate of 4 classes for different algorithms

Fig. 10. Accuracy, Precision, Recall, F1 Score, MCC, and CSI rate Comparison Bar Chart for Individual ML Models.

These performance scores are listed according to the well-nourished, mildly nourished, moderately nourished, and severely malnourished sequel, which is also true for each confusion matrix, and each algorithm was trained using 80% of the dataset and 20% for testing.

For the statistical analysis in Table 4, a total of 10 features were incorporated, including the targeted feature "Mstatus", which is the four classes of malnutrition stages. According to this Pearson correlation coefficient, nine of the 18 features have a negatively significant correlation with the increasing stages of malnutrition. It indicates WAZ2, WHZ2, ZBMI, BD3, HAZ2, CA1, Windex5, melevel, and CA14/CA16/CA17 have direct impacts in ascending order on causing a child to be malnourished. So, the more negative the features, the more malnourished a child would be.

Comparing both ML (RF) and statistical approaches, a strong dependency risk for a child was the same for WAZ2, WHZ2, ZBMI, HAZ2, BD3, Windex5, and melevel in terms of significance. That signifies the performance success in both approaches, but the Pearson correlation detected CA1 and CA14/CA16/CA17, which are "whether the child had diarrhea in the last 2 weeks" and "whether the child had illness with fever, cough, and breathing," respectively; those are missing in the RF classifier. On the other hand, UB2, HH7, HL4, and CA22/CA23L/CA23M/CA23N are the "age of the child.". The RF algorithm could detect the "division of the respondent," "gender of the child," and "whether the child had medication during cough," but the Pearson correlation was unable to.

**Table 6**
Performance comparison with the existing literature

| No | Study | The best-fitted model | Dataset | Result (%) |
|---|---|---|---|---|
| 1. | Talukdar et al. [5] | RF | BDHS 2014 | Accuracy 68.51% |
| 2. | Shahriar et al. [6] | ANN | BDHS 2014 | Accuracy 74.40% (Avg.) |
| 3. | Rahman et al. [7] | RF | BDHS 2014 | Accuracy 87.23% (Avg.) |
| 4. | Mansur et al. [8] | BLR | BDHS 2014 | Accuracy 67.40% (10-fold CV) |
| 5. | Khan et al. [9] | GBOOST | BDHS 2014 | Accuracy 68.47% |
| 6. | Hemo et al. [10] | RF | BDHS 2014 | Accuracy 71.25% (Avg.) |
| 7. | This study | MLP | MICS 2019 | Accuracy 98.88% |

Table 6. depicts a performance comparison with some of the existing literature that has been studied, considering the malnourished children of Bangladesh using machine learning algorithms. Our study surpassed the degree of accuracy obtained in other analyses, and in this case, MLP was found to be the best-performing model.

**Limitations and future works**
Though a detailed analysis from the MICS – 2019 dataset was done in this work, it has some limitations that can be considered. Firstly, the dataset contained so many features that it was very difficult to include all of them separately. To address this issue, the relevant features were merged and analyzed in this work. Future research could incorporate those additional features individually to obtain any other significant correlation. Secondly, supervised machine learning algorithms were used in this work to classify different malnutrition status and identify the risk factors. However, future research could consider unsupervised machine learning algorithms for clustering the features to recognize any significant pattern in the data.

**Conclusion**
Persisting malnutrition is a burdening concern in developing countries with a colossal health risk. In our study, seven of the training set features were found to be significant in both approaches, and other less significant components were also shown. Both the RF and MLP algorithms performed incredibly well when the RF executed the weighted classes, with MLP slightly outperforming the RF for the classification of the four classes. This study would help to identify the children at risk for malnutrition, and general guidelines can be followed by policymakers, NGOs, and healthcare service providers to address the identified risk factors. Targeting and working on the vital impact creator components will ease prevention and cure the burdening issue in our society.

**Data availability statement**

This work has been done based on the Multiple Indicator Cluster Survey (MICS – 2019) dataset available upon request on the website of Unicef [22].